\documentclass{article}

\usepackage{arxiv}

\usepackage[utf8]{inputenc} % allow utf-8 input
\usepackage[T1]{fontenc}    % use 8-bit T1 fonts
\usepackage{hyperref}       % hyperlinks
\usepackage{url}            % simple URL typesetting
\usepackage{booktabs}       % professional-quality tables
\usepackage{amsfonts}       % blackboard math symbols
\usepackage{nicefrac}       % compact symbols for 1/2, etc.
\usepackage{microtype}      % microtypography
\usepackage{lipsum}		% Can be removed after putting your text content
\usepackage{graphicx}
\usepackage{natbib}
\usepackage{doi}
\usepackage{array}
\usepackage{multirow}
\usepackage{float}
\usepackage{subfigure}
\usepackage{amsmath,amssymb,amsfonts}
\usepackage{algorithmic}
\usepackage{graphicx}
\usepackage{textcomp}
\usepackage{array}
\usepackage{times}
\usepackage{epsfig}
\usepackage{graphicx}
\usepackage{amsmath}
\usepackage{amssymb}
\usepackage{booktabs}
\usepackage{threeparttable}
\usepackage{bm}
\usepackage{multirow}
\usepackage{float}
\usepackage{subfigure}
\usepackage{indentfirst}
\usepackage{amsmath}

\title{Research on Gender-related Fingerprint Features}

\author{ \href{https://orcid.org/0000-0002-4930-5978}{\includegraphics[scale=0.06]{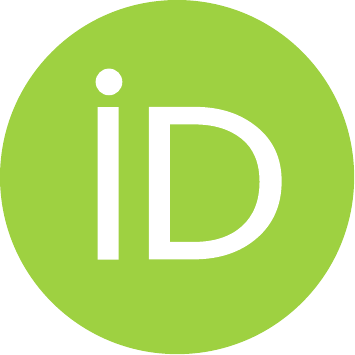}\hspace{1mm}Yong Qi}\thanks{Corresponding Author}\hspace{1.8mm}\text{\textsuperscript{1, 2}} \\
	\texttt{qiyong@sust.edu.cn} \\
	\And
	\href{https://orcid.org/0000-0003-4469-0173}{\includegraphics[scale=0.06]{orcid.pdf}\hspace{1mm}Yanping Li}\hspace{0.5mm}\text{\textsuperscript{1, 2}}\\
	\texttt{yanpinglics@gmail.com} \\
	\And
	\href{https://orcid.org/0000-0002-2965-3158}{\includegraphics[scale=0.06]{orcid.pdf}\hspace{1mm}Huawei Lin}\hspace{0.5mm}\text{\textsuperscript{1, 2, 3}}\\
	\texttt{huaweilin.cs@gmail.com} \\
	\And
	\href{https://orcid.org/0000-0002-4916-3025}{\includegraphics[scale=0.06]{orcid.pdf}\hspace{1mm}Jiashu Chen}\hspace{0.5mm}\text{\textsuperscript{1, 2}}\\
	\texttt{gaasyu.chan@gmail.com} \\
	\And
	{\hspace{1mm}Huaiguang Lei}\hspace{0.5mm}\text{\textsuperscript{1}}\\
	\texttt{leihg@sust.edu.cn}
}

\date{}

\hypersetup{
	pdftitle={Research on Gender-related Fingerprint Features},
	pdfsubject={cs.AI},
	pdfauthor={Yong Qi, Yanping Li, Huawei Lin, Jiashu Chen, Huaiguang Lei},
	pdfkeywords={Fingerprint, Gender Identification, Finger Contribution, Feature Visualization},
}

\begin{document}
	\maketitle
	
	\begin{abstract}
		Fingerprint is an important biological feature of human body, which contains abundant gender information. At present, the academic research of fingerprint gender characteristics is generally at the level of understanding, while the standardization research is quite limited. In this work, we propose a more robust method, Dense Dilated Convolution ResNet (DDC-ResNet) to extract valid gender information from fingerprints. By replacing the normal convolution operations with the atrous convolution in the backbone, prior knowledge is provided to keep the edge details and the global reception field can be extended. We explored the results in 3 ways: 1) The efficiency of the DDC-ResNet. 6 typical methods of automatic feature extraction coupling with 9 mainstream classifiers are evaluated in our dataset with fair implementation details. Experimental results demonstrate that the combination of our approach outperforms other combinations in terms of average accuracy and separate-gender accuracy. It reaches 96.5\% for average and 0.9752 (males) /0.9548 (females) for separate-gender accuracy. 2) The effect of fingers. It is found that the best performance of classifying gender with separate fingers is achieved by the right ring finger. 3) The effect of specific features. Based on the observations of the concentrations of fingerprints visualized by our approach, it can be inferred that loops and whorls (level 1), bifurcations (level 2), as well as line shapes (level 3) are connected with gender. Finally, we will open source the dataset that contains 6000 fingerprint images.\let\thefootnote\relax\footnotetext{\hspace{-1.5mm}1. School of Electronic Information \& Artificial Intelligence, Shaanxi University of Science \& Technology, Xi'an 710021, China}\let\thefootnote\relax\footnotetext{\hspace{-1.5mm}2. Shaanxi Joint Laboratory of Artificial Intelligence (Shaanxi University of Science \& Technology) , Xi'an 710021, China}\let\thefootnote\relax\footnotetext{\hspace{-1.5mm}3. College of Computer Science and Software Engineering, Shenzhen University, Shenzhen 518060, China}
	\end{abstract}

	% keywords can be removed

	\section{Introduction}
	Fingerprint gender identification aims to extract gender-related features from an unidentified fingerprint to recognize one's gender information. It can be divided into two stages, namely extracting as well as classifying~\cite{abdullah2016fingerprint,abdullah2016support,gnanasivam2012fingerprint,gupta2014fingerprint,mishra2017novel,rekha2019dactyloscopy,shinde2015analysis,wedpathak2018fingerprint}, in which the former step is of great significance since the effectiveness of gender identification, is primarily determined by the sufficiency of gender-related features. Nowadays, Fingerprints can be classified into three level\cite{karu1996fingerprint}\cite{henry1913classification} as shown Figure 1. Classifying ridge-related features extracted manually has achieved fairly good results, reaching an overall accuracy for 90\% for average\cite{arun2011machine,badawi2006fingerprint,wedpathak2018fingerprint}. High performances, however, depends strongly on the manual extraction of features from well-selected regions\cite{kralik2003epidermal}. These methods have major shortcomings, such as high error, weak robustness, and high labor consumption. So far, with the growing popularity of machine learning and deep learning, automatic feature extraction has become a major foucus.\\% necessary to decrease error, liberate manpower as well as improve robustness.\\
	
	\begin{figure}[h!]
		\centering %图片居中
		\includegraphics[width=1\textwidth]{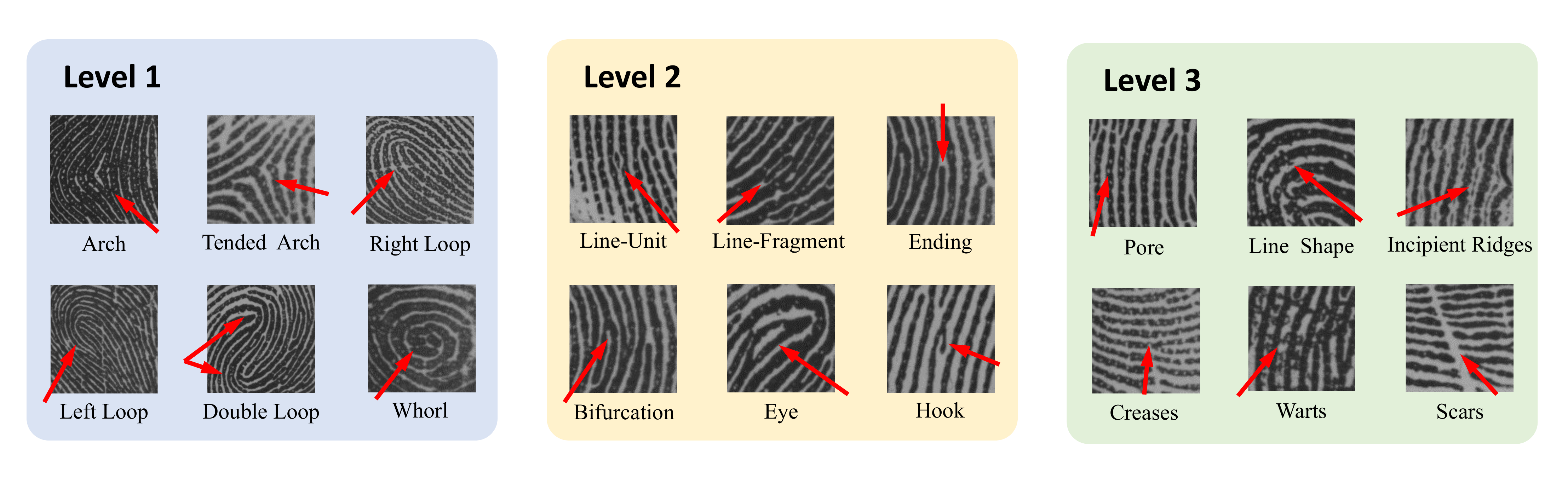}
		\caption{Three levels of fingerprints} %最终文档中希望显示的图片标题
		\label{fig:fig1}
	\end{figure}
	
	\begin{figure}[h!]
		\centering
		\includegraphics[width=0.8\textwidth]{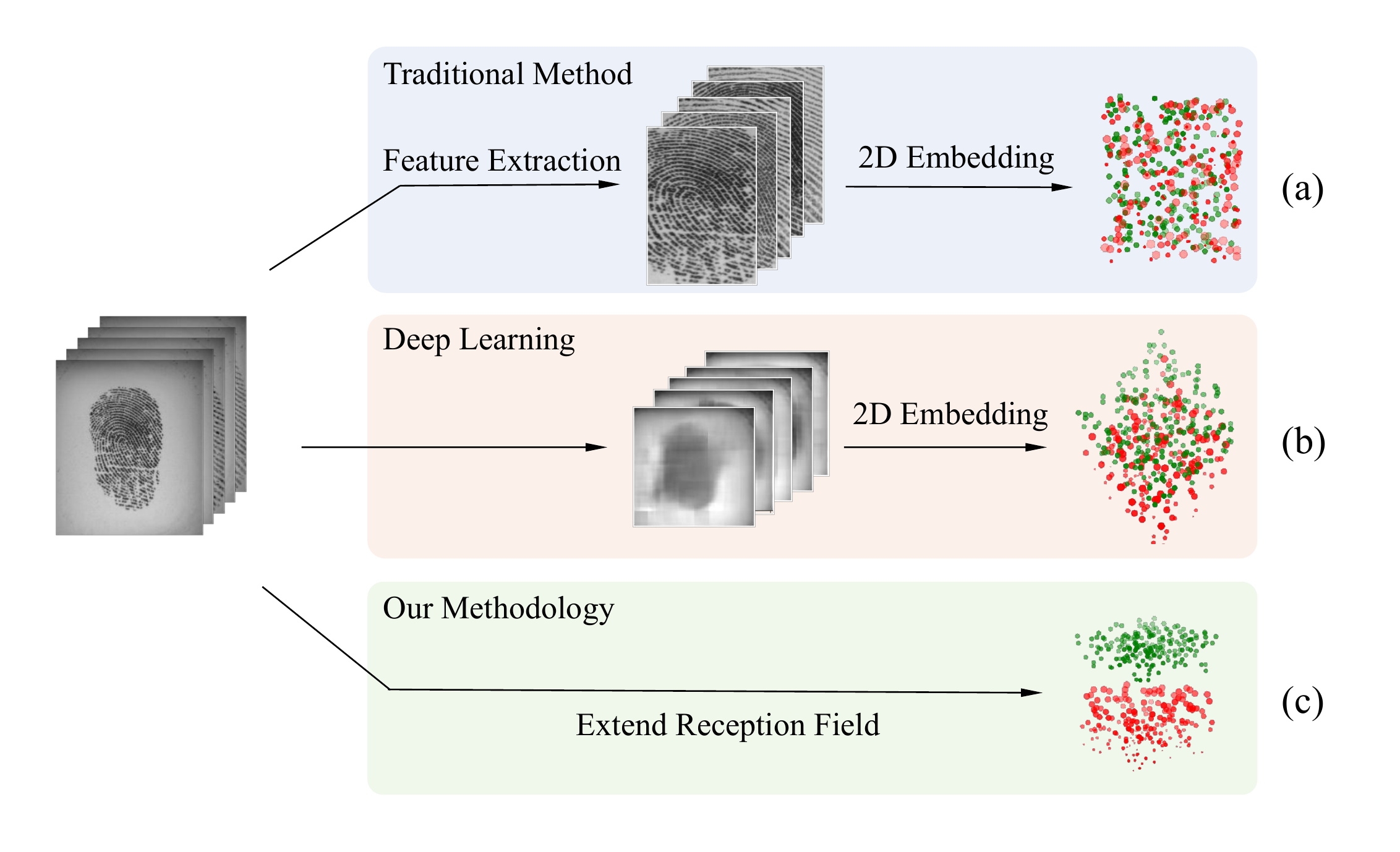}
		\caption{General automatic gender classification methods preview. (a) shows embedding and feature extracted by traditional method, which are difficult to distinguish; (b) shows deep learning method, and it is easier to identify than (a); and (c) is the result of our experiment and it performs best.} 
		\label{fig:fig2}
	\end{figure}
	
	To realize the automatic feature extraction, considerable work has been done. In machine learning, methods such as DWT, SVD, PCA as well as FFT are extensively used \cite{gnanasivam2012fingerprint,kaur2012fingerprint,marasco2014exploiting}. For deep learning based methods, including deep autoencoder neural networks such as VGG and ResNet,etc have been investigated \cite{chen2017deep}.\\
	
	Although numerous algorithms have been proposed, there are still 2 major challenges. First, traditional approaches aquire excessive labor consumption and lack automation. Second, the automatic feature extraction method lacks robustness, which only concerns regions instead of considering the global field. Figure 2 shows the difference between the normal automatic feature extraction method and the deep neural network with global reception consideration. The latter can divide the feature space normatively. Besides, private datasets with disparate data distributions and sizes will directly influence the accuracy. Thus, because of the above challenges, we propose a global feature extraction method to improve the efficiency of gender classification.\\
	
	\textbf{Contributions of this paper:}\\
	\hspace*{0.6cm}1. We propose a feature extraction method of fingerprint that takes global features into account. For existing typical automatic feature extraction methods and classification methods, we make comprehensive comparisons with fair implementation details.\\
	\hspace*{0.6cm}2. We make a comprehensive comparison to test the efficiency of our method. The finger with the richest gender-related features is detected using the highest performance method. Finally, we visualize the concentrations using the selected method, which indicates the regions with the highest contributions and their corresponding specific features in the gender identification task.\\
	\hspace*{0.6cm}3. We will finally open source the dataset since the open-source datasets in fingerprint gender classification are limited.\\
	
	The following article will be divided into 4 parts. Part 2 summarizes the development over the last few years. The introduction to automatic feature extraction and classification methods used in this article is provided in part 3. All results and discussions can be found in part 4. At last, we will conclude and outlook future work.

	%--------------------------------%
	\section{Related Work}
	Previous discussions have demonstrated that gender-related features have a significant impact on fingerprint gender identification. In this section, we will first go through the evolution and then review the progress in recent years in a conclusive table, which is shown in Table~\ref{pre_methods}.\\
	
	In 1999, Acree\cite{acree1999there} manually counted ridges in specific areas on the fingerprint epidermis, showing that ridge count can determine the gender. Then in 2003, the mean epidermal ridge breadth has been proposed to identify gender~\cite{kralik2003epidermal}. Similarly in 2006, Badawi~\cite{badawi2006fingerprint} manualy extracted ridge counts, ridge thickness to valley thickness ratio(RTVTR), and white lines count to determine the gender using a neural network as a classifier. Later, Gungadin~\cite{gungadin2007sex} found a threshold of the ridge density 13ridges/25mm$^{2}$ in 2007, which determines the gender as male when the ridge density is lower than the threshold. In other words, the female has higher ridge density is probably due to they have lower ridge breadth~\cite{kralik2003epidermal}. \\
	%Automated feature extraction 
	
	Owing to the inevitable error and high manpower consumption of manual feature extractions, automatic fingerprint extraction has been proposed. In 2012, ridge features have been analyzed in the spatial domain using FFT, 2D-DCTT, and PSD~\cite{kaur2012fingerprint}, reaching an accuracy of 90\% for females and 79.07\% for males, respectively. In the same year, Gnanasivam~\cite{gnanasivam2012fingerprint} proposed a method using DWT and SVD as the feature extractor. In 2014, Marasco~\cite{marasco2014exploiting} utilized LBP and LPQ on texture features and used PCA to reduce the features, then the kNN classifier was applied to the extracted features. In 2017, DWT feature extraction was applied, matching with neural network~\cite{gupta2014fingerprint}. Since the presence of the deep convolution autoencoder neural network, deep learning has been widely applied in the feature extraction task especially in the biometric recognition field~\cite{chen2017deep}.
	
	%------------------------------------Literature Survey Table
	\begin{table*}[htbp]
		\renewcommand\arraystretch{1}
		\caption{\newline The overview of the previous methods.}
		\centering
		\begin{center}
			\begin{tabular}{|m{1cm}<{\centering}|m{2.5cm}<{\centering}|m{3cm}<{\centering}|m{1.5cm}<{\centering}|m{3cm}<{\centering}|m{2cm}<{\centering}|}
				%\toprule  %添加表格头部粗线
				\hline
				\textbf{Year}& \textbf{Publisher}& \textbf{Feature Extraction}& \textbf{Classifier}& \textbf{Results}& \textbf{Dataset}\\
				%\midrule  %添加表格中横线
				\hline
				1999& Acree~\cite{acree1999there}& Ridge counting manually& Threshold& Female ridge density is higher& Own dataset contains 400 subjects\\
				\hline
				2003& Kralik~\cite{kralik2003epidermal}& Mean epidermal ridge breadth& Threshold& Ridge
				breadth is 9\% greater in males than in females& Own dataset contains 60 subjects\\
				\hline
				2006& Badawi~\cite{badawi2006fingerprint}& RTVTR, white line, ridge thickness & Neural Network& Average 88.8\%& Own dataset contains 220 subjects\\
				\hline
				2007& Gungadin~\cite{gungadin2007sex}& Counting ridges in the upper portion of the radial border& Threshold& Ridge density of male's fingerprint tend to be less than or equal to 13ridges/25mm$^{2}$& Own dataset contains 500 subjects\\
				\hline
				2011& Arun~\cite{arun2011machine}&  Ridge count, ridge density, white line and RTVTR& SVM with RBF kernel& Overall 96\%& Own dataset contains 150 male and 125 female images\\
				\hline
				2012& Kaur~\cite{kaur2012fingerprint}& FFT, DCT, PSD& Threshold& 90\% for female \& 79.07\% for male & Own dataset contains 220 subjects\\
				\hline	
				2012& Gnanasivam~\cite{gnanasivam2012fingerprint}& DWT+SVD& KNN&  91.67\% for male \& 84.69\% for female &Own dataset contains 357 subjects\\
				\hline
				2014& Marasco~\cite{marasco2014exploiting}& LBP and LPQ descriptor & KNN& Overall 88.7\%& Own dataset contains 494 subjects\\
				\hline
				2014& Gupta~\cite{gupta2014fingerprint}& DWT& ANN& Overall 91.45\% & Own dataset contains 55 subjects\\
				\hline
				2016& Abdullah~\cite{abdullah2016fingerprint}& Ridge count, ridge density, white line and RTVTR& J48& Overall 96.28\%& Own dataset contains 296 subjects\\
				\hline
				2016& Abdullah~\cite{abdullah2016multilayer}& Ridge count, ridge density, white line and RTVTR& MLP& Overall 97.25\%& Own dataset contains 300 subjects\\
				\hline
				2017& Sheetlani~\cite{sheetlani2017fingerprint}& DWT& CNN& Overall 96.60\%& Own dataset contains 80 subjects\\
				\hline
				2017& Ashish Mishra~\cite{mishra2017novel}& minutiae, incipient ridges& SVM \& NN& 76.06\% for SVM and 83.7
				\% for female & NIST\\
				\hline
				2018& Wedpathak~\cite{wedpathak2018fingerprint}& Ridge count \& RTVTR & ANN & 88\% for male \& 78\% for female& Own dataset\\
				\hline
				2019& Alam~\cite{alam2019comparative}& DWT+SVD & KNN& 91.25\% for male \& 88.96\% for female& Own dataset contains 42 subjects\\
				\hline
				2019& Rekha~\cite{rekha2019dactyloscopy}& Gabor filter& KNN, SVM, Naive Bayes& $\backslash$ & $\backslash$ \\
				\hline
				2020& DDC-ResNet (Ours)& Autoencoder& CNN& \textbf{97.52\%} for male \& \textbf{95.48\%} for female  & Dataset contains 200 subjects \\
				\hline
				%\bottomrule %添加表格底部粗线
			\end{tabular}
		\end{center}
		\label{pre_methods}
	\end{table*}
	
	%------------------------------------End Literature Survey Table
	\section{Overview of Feature Extraction and Classification Algorithms}
	In this section, we will introduce our approaches and the implementation principles of automatic feature extraction methods. In addition, we will outline the classification methods utilized in this paper.
	
	\subsection{Feature Extraction}
	\subsubsection{Discrete Wavelet Transformation}
	Wavelet has been extensively applied in feature extraction, soft-biometrics recognition, and denoising, etc. It decomposes an image into sub-bands containing frequency and orientation information to represent the valid signals. Specifically, a fingerprint image is decomposed into 4 sub-bands at one level, namely low-low (LL), low-high (LH), high-low (HL) and high-high (HH) which is shown in Figure 3. Typically, the LL sub-band will be decomposed repeatedly since it is thought to represent the most energy, and k refers to the repeat times. If k is set, $(3\times k)+1$ sub-bands are available. The energy of each sub-band is calculated by equation~\ref{con:dwt}, which will be used as a feature vector for gender classification ($E_{k}$), where $X_{k}(i,j)$ represents the pixel at the position $i$ and $j$ on the $k-th$ level. $W$ and $H$ represent the width and the height of the sub-band, respectively.
	
	%------------------------------dwt formula
	\begin{equation}
		E_{k} = \frac{1}{WH}\sum_{i=1}^{W}\sum_{j=1}^{H}{|X_{k}(i,j)|}
		\label{con:dwt}
	\end{equation}
	%------------------------------End dwt formula
	
	%---------------------------DWT Figure-----------------------------------------
	
	\begin{figure}[h] 
		\centering 
		\includegraphics[width=8.5cm]{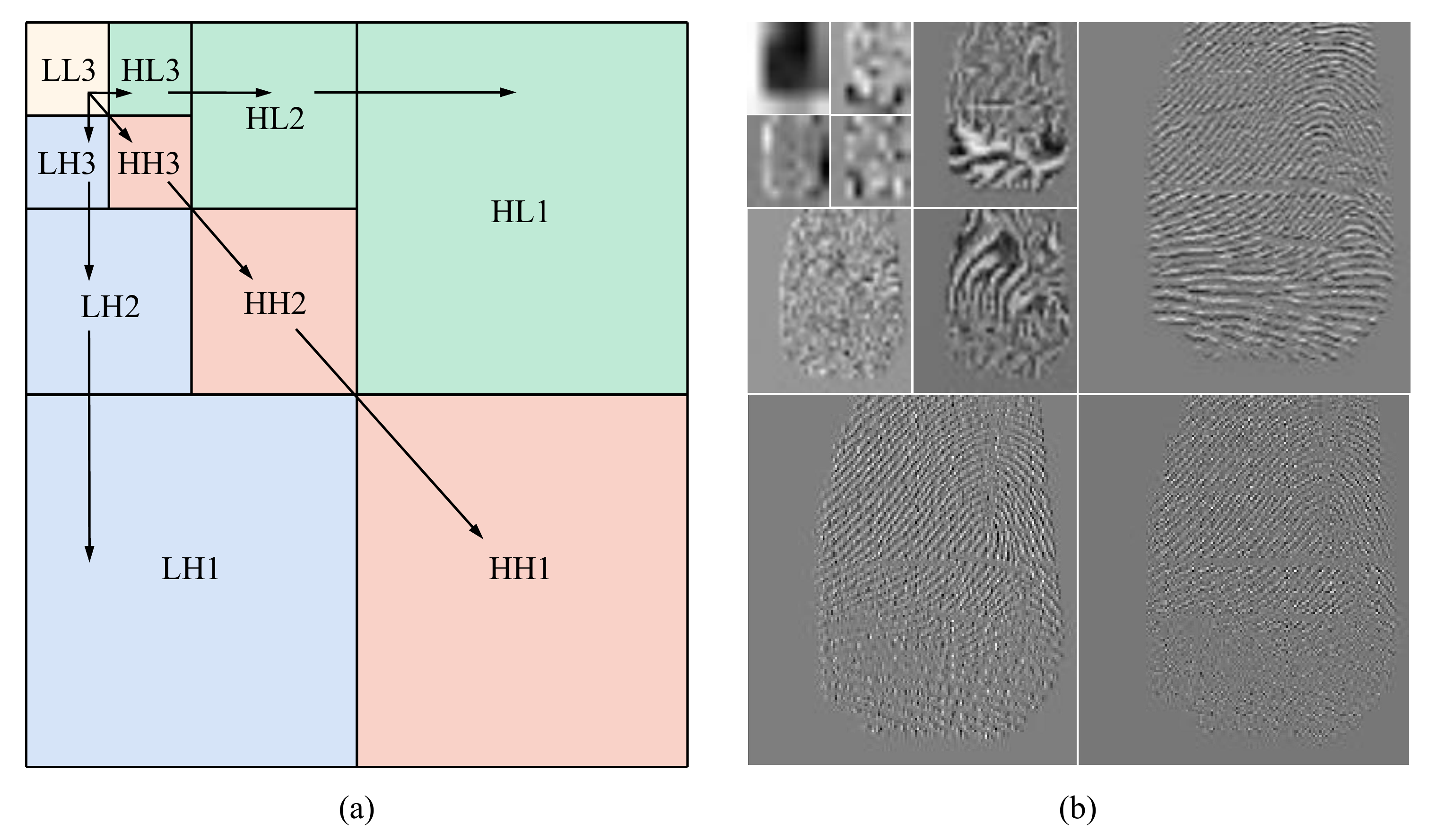} 
		\caption{Discrete Wavelet Transformation (DWT). (a) Representation of level 3 DWT. (b) Fingerprint at different 3-level DWT sub-bands}
		\label{Intro_fig}
	\end{figure}
	%%%%%%%%%%%%%%% End Test
	%--------------------------------------
	
	%--------------------- End DWT Figure-----------------------------------------
	\subsubsection{Singular Value Decomposition}
	The fundamental of the SVD is that any rectangular matrix can be transformed into the product of three new matrices. Specifically, given a fingerprint image matrix $A$ with $H$ rows and $W$ columns, it can be factored into $U$, $S$, and $V$ by using equation~\ref{con:ksvd}, where $U = AA^{T}$ and $V = A^{T}A$ and S are diagonal matrices that contain the square root eigenvalues with the size of $H$ by $W$, which is stored for gender classification.
	
	%------------------------------ksvd formula
	\begin{equation}
		A = U S V^{T}
		\label{con:ksvd}
	\end{equation}
	
	\subsubsection{Fast Fourier Transform}
	The FFT is used to transform a fingerprint image into the frequency domain. The transformed vector contains most of the information in the spatial domain and is used for gender classification. It is presented in equation ~\ref{con:fft1}, where M and N represent the height and width of the fingerprint image, k and l represent frequency variables, respectively. $0 \leq m$, $k \leq M-1$, $0 \leq n$, $1 \leq N-1$.
	%------------------------------fft formula
	
	\begin{equation}
		F[k,l] = \frac{1}{\sqrt{MN}}\sum_{n=0}^{N-1}\sum_{m=0}^{M-1}f[m,n]e^{-j2\pi(\frac{mk}{M}+\frac{nl}{N})}
		\label{con:fft1}
	\end{equation}
	
	%--------------------------------------
	\subsubsection{Our method}
\begin{figure}[h!] 
		\centering 
		\includegraphics[width=9cm]{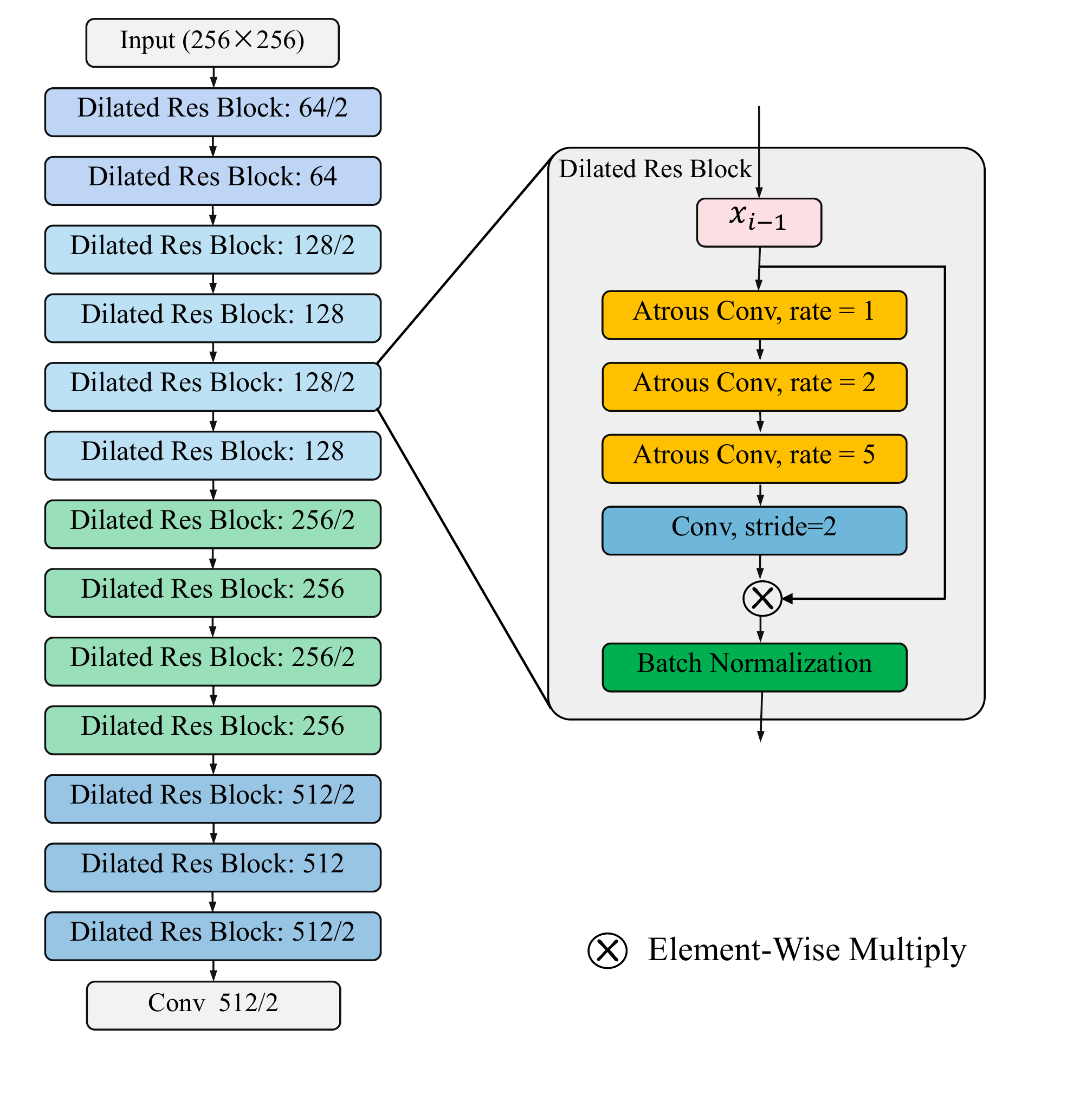} 
		\caption{Architecture of DDC-ResNet for gender information extracting. The input fingerprint image is of size $256\times 256$. (Right) The basic dilated residual block in the DDC-ResNet, where $x_{i-1}$ is the feature map from previous block.}
		\label{model_fig}
	\end{figure}

	Autoencoder neural networks can be divided into the encoder and decoder. Concretely, in the encoder step, the low dimensional data will be compressed into feature vector in high dimensional space, and the decoder step is to reconstructs the original data without redundant features. The general equation is described in ~\ref{con:AE}. The feature vector can be optimized by minimizing the distance between the original data and the reconstructed data, and used for gender classification. Our method utilizes ResNet as the backbone, and uses the atrous convolution operation to replace the normal convolutions, as shown in Figure \ref{model_fig}. In the block res we utilize atrous rates as 1, 2 and 5 to prevent the gridding effect.
	
	%-------------------------------AE formula
	\begin{equation}
		y' = D(E(x),x)
		\label{con:AE}
	\end{equation}
	
	%-------------------------------End AE formula
	% + say more
	%-------------------------------model
	
	%-----------------------------End model
	%-----------------------------End model
	\subsection{Gender Classification}
	To ensure fairness of experiments, we adopt commonly used 9 classifiers in gender classification problem, which are CNN, SVM (3 kernels), kNN, Adaboost, J48, ID3, and LDA. Among them the 3 kernels in SVM refer to linear, radial basis function, and polynomial, respectively. CNN is comprised of fully connected layers. These are all mainstream algorithms used in classification tasks. We threrfore only highlight their implementation details in the next section and more details of the methods are presented in section 4.1.
	%-------------------------------------------------------------------------
	%-------------------------------General Preview
	%\begin{figure}[htbp] 
	%\centering 
	%\includegraphics[height=3.5cm, width=6cm]{general_preview.pdf} 
	%\caption{General automatic gender classification methods preview}
	%\label{fig:general_preview}
	%\end{figure}
	%-----------------------------End General Preview
	
	%%%%%%%%%%%%%%%%%%%%%%%%%%%%%%%%%%%%%%%%%%EXPERIMENTAL RESULTS%%%%%%%%%%%%%%%%%%%%%%%%%%%%%%%%%%%%%%%%%%%%%%%%%%
	
	\section{Experimental Results}
	In this section, we first introduce our dataset and adopted implementation details. In the experimental stage, factors that affect gender-related features will be analyzed comprehensively to provide some useful conclusions.%3 subsections are partitioned to research on the gender-related features. Gender-related features has been researched on 3 aspects, which are .
	
	\subsection{Experimental Setup}
	%+we acquisite the same fingerprint for twice because...
	\subsubsection{Dataset}
	The fingerprint dataset used in experiments is obtained from ZK fingerprint acquisition equipment with 500 dpi, containing 200 persons (102 females and 98 males) and 6000 images. Each finger is collected 3 times to guarantee the quality of the fingerprint image.
	\subsubsection{Implementation Details}
	%gradation
	In the preprocessing step, we resize each fingerprint image to $256\times 256$ and normalize it to $[0,1]$. $Train:Test$ ratio is $4:1$, No repetitive finger is guaranteed in both the training set and test set.\\
	
	In the feature extraction stage, for DWT, each fingerprint goes through eight levels of decomposition. For SVD, each fingerprint image vector is of size 256. In VGG-Net, we apply the VGG-19 network in which 8 blocks are utilized. Each block contains 2 convolutional layers and a batch of normalization layers. ResNet-18 is applied and each block contains a residual block to prevent the vanishment of gradients. The DDC-ResNet, by replacing the normal convolutional layers with dilated convolutional layers in ResNet, lowers the loss of valid edge features~\cite{wang2018understanding}. To keep the fairness, these three types of feature extraction networks are of the vector size $[-1,512]$ in the last layer. Here we set the batch size to be 10, and the iteration numbers are 10K. The optimizer we adopt is Adam, in which the learning rate is $3 \times 10^{-4}$. Inspired by Maas~\cite{maas2013rectifier}, the leaky relu is used as the activation function. All the codes are realized with the framework TensorFlow~\cite{abadi2016tensorflow} in python. Hardware utilization includes GPU with NVIDIA TITAN Xp, and CPU with 2.8GHz, 32GB in RAM.\\ 
	
	%Classification settings
	In the classification progress, the Adaboost is used with the $n\_estimator = 100$, $learning$ $rate = 0.01$, $max\_depth = 7$, and $subsample = 1.0$. In SVM, the linear kernel is utilized with the function LinearSVC in the default configuration, $C = 100$ and $gamma = 0.5$ are set when applying RBF and polynomial kernels. The parameter k in kNN is set to be 1. J48 algorithm is conducted in WEKA software with the default settings~\cite{holmes1994weka}. The CNN classifier is comprised of 3 fully connected layers with the leaky relu activation. Classification algorithms except J48 and CNN are all conducted in scikit-learn~\cite{pedregosa2011scikit}.\\
	
	%实验目的，验证说明了什么问题-》做的什么实验，具体所用的方法/数据集-》实验结果在哪里-》实验现象，表格中明显数值-》实验结果
	%%%%%%%%%%%%%%%%%%%%%%%%%%%%%table
	\begin{table}[htbp]
		\renewcommand\arraystretch{2}
		\caption{The upper value in each cell represents $average\ accuracy$ (\%), and the under value shows separate gender accuracy, meaning $male\ accuracy/female\ accuracy$ .}
		\centering
		\begin{center}
			\resizebox{\textwidth}{!}{
				\begin{tabular}{|p{2cm}<{\centering}|c|c|c|c|c|c|c|c|c|}
					\hline  %添加表格头部粗线
					& \textbf{Adaboost}& \textbf{SVM$^{1}$}& \textbf{SVM$^{2}$}& \textbf{SVM$^{3}$}& \textbf{KNN}& \textbf{J48}& \textbf{CNN}& \textbf{ID3}& \textbf{LDA}\\
					%\midrule  %添加表格中横线
					\hline
					\multirow {2}{1.5cm}{\centering\textbf{FFT}}
					& 85.280& 65.827& 82.604& 87.310& 92.202& 73.912& 87.220& 78.931& 88.498 \\
					& 0.8337/0.8719& 0.6895/0.6270& 0.7935/0.8586& 0.8653/0.8809& 0.9346/0.9094& 0.7224/0.7558& 0.8814/0.8630& 0.7682/0.8104& 0.9012/0.8687\\
					\cline{1-10}
					\multirow {2}{1.5cm}{\centering\textbf{DWT}}
					& 91.494& 70.001& 88.275& 88.735& 90.460& 90.394& 91.360& 90.344& 87.192\\
					& 0.9531/0.8768& 0.7280/0.6720& 0.9144/0.8511& 0.9452/0.8295& 0.9187/0.8905& 0.8977/0.9102& 0.9455/0.8817& 0.8727/0.9342& 0.8693/0.8745\\
					\cline{1-10}
					\multirow {2}{1.5cm}{\centering\textbf{SVD}}
					& 92.096& 66.552& 86.506& 91.609& 88.230& 76.667& 92.413& 89.770& 89.721\\
					& 0.9294/0.9125& 0.6839/0.6471& 0.7910/0.9391& 0.9248/0.9074& 0.8917/0.8729& 0.7296/0.8037& 0.9350/0.9133& 0.8856/0.9098& 0.9411/0.8443\\
					\cline{1-10}
					\multirow {2}{1.5cm}{\centering\textbf{ResNet}}
					& 92.068& 69.302& 85.502& 91.713& 92.529& 90.460& 93.333& 90.460& 82.106\\
					& 0.9334/0.9080& 0.7202/0.6658& 0.8210/0.8890& 0.9220/0.9123& 0.9072/0.9434& 0.8979/0.9113& 0.9463/0.9204& 0.9149/0.8943& 0.7966/0.8455\\
					\cline{1-10}
					\multirow {2}{1.5cm}{\centering\textbf{VGG}}
					& 94.138& 72.454& 88.506& 94.713& 93.563& 88.220& 93.678& 90.459& 89.385\\	
					& 0.9464/0.9364& 0.7208/0.7283& 0.8192/0.9509& 0.9590/0.9353& 0.9428/0.9285& 0.9195/0.8449& 0.9423/0.9313& 0.9099/0.8993& 0.9047/0.8830\\	
					\cline{1-10}
					\multirow {2}{1.5cm}{\centering\textbf{DDCResNet}}
					& 94.253& 73.844& 89.576& \textbf{95.333}& 92.759& 89.344& \textbf{96.500}& 91.494& 93.599\\
					& 0.9350/0.9501& 0.7366/0.7403& 0.8260/0.9655& \textbf{0.9743/0.9324}& 0.8865/0.9687& 0.8921/0.8948& \textbf{0.9752/0.9548}& 0.9159/0.9140& 0.9461/0.9259\\
					\cline{1-10}
					%\bottomrule %添加表格底部粗线
			\end{tabular}}
		\end{center}
		\label{tb1}
	\end{table}
	
	%%%%%%%%%%%%%%%%%%%%%%%%%%% END table
	\subsection{Effect of Methods}
	To evaluate the effect of different methods on gender-related features, we make a benchmark for the combinations of 6 feature extraction methods coupling with 9 classifiers to compare their average and separate-gender accuracies, as shown in Table~\ref{tb1}. In terms of the average accuracy, the DDC-ResNet extractor outperforms other extraction methods when matching with different classifiers. For the classifiers, the Adaboost, SVM with the polynomial kernel, and CNN are more outstanding. On the contrary, the SVM with the linear kernel seems not qualified for the extracted features with high dimensional space in the gender identification task. Regarding the separate-gender performance, most of the results show more correct predictions in males than in females, especially for DWT+SVM$^{3}$. The difference is more than 10\%, which is agreed with the\cite{shinde2015analysis,gnanasivam2019gender}, suggesting that fingerprints of males contain richer gender-related features in some aspects. To make a more comprehensive analysis, we also evaluate the time consumption of each combination when feeding different batch sizes of data. The result is shown in Table~\ref{TimeMeasurement}. In general, deep learning extractors consume more time than machine learning, and SVD extractor consumes the least time. However, FFT behaves oppositely which is because the transformation does not reduce the feature size. In summary, the combination of DDC-ResNet and CNN outperforms other combinations, reaching an average accuracy of 96.50\% and separate-gender accuracy of 0.9743/0.9324. Moreover, males contain richer gender-related features than females. After evaluating the performance of various methods with all test fingers, we will explore how specific fingers influence the gender identification results below.
	\subsection{Effect of Fingers}
	After studying the effect of varing methods on gender-related features, we further explore the effect of each finger. We divide the testing fingerprints into 10 sets, each set of which corresponds to a specific finger containing 660 fingerprint images. We apply the DDC-ResNet coupled with CNN (best performance method above) to test the 10 sets, as listed in Table~\ref{bestfinger}. The result indicates that for each specific finger, the right ring finger (R2) shows the highest accuracy, reaching 92.455\%. For 5 pairs of fingers, ring fingers outperform other pairs, reaching an accuracy of 91.413\%. For the overall hand, the right hand achieves a higher accuracy than the left hand, which reaches 87.872\%. To better understand the effect of fingers on gender-related features, more careful studies will be carried out in the following part.\\
	%%%%%%%%%%%%%%%%%%%%%%%%%%% table3
	
	\begin{table}[h]
		\renewcommand\arraystretch{1}
		\caption{Estimate performance of each finger (\%), in which L means left hand and R means right hand, $F_{x}$ represents Finger $x$. The index from 1 to 5 means little finger, ring finger, middle finger index finger and thumb respectively.}
		\centering
		\begin{center}
			\begin{tabular}{|p{0.5cm}<\centering|p{1cm}<\centering|p{2cm}<{\centering}|p{2.5cm}<{\centering}|}
				\hline  %添加表格头部粗线
				L1& 88.291& \multirow{2}{2.5cm}{\centering $F_{1}$: 87.078}& \multirow{5}{2.5cm}{\centering $L_{average}$: 86.418}\\
				\cline{1-2}
				R1& 85.864& & \\
				\cline{1-3}
				L2& 90.371& \multirow{2}{2.5cm}{\centering $F_{2}$: \textbf{91.413}}& \\
				\cline{1-2}
				R2& \textbf{92.455} & & \\
				\cline{1-3}
				L3& 87.152& \multirow{2}{2.5cm}{\centering $F_{3}$: 89.258}& \\
				\cline{1-2}
				\cline{4-4}
				R3& 91.363 & & \multirow{5}{2.5cm}{\centering $R_{average}$: \textbf{87.872}} \\
				\cline{1-3}
				L4& 82.819& \multirow{2}{2.5cm}{\centering $F_{4}$: 84.637}& \\
				\cline{1-2}
				R4& 86.455 & & \\
				\cline{1-3}
				L5& 83.455& \multirow{2}{2.5cm}{\centering $F_{5}$: 83.340}& \\
				\cline{1-2}
				R5& 83.224 & & \\
				\hline
			\end{tabular}
			\label{bestfinger}
		\end{center}
	\end{table}
	%%%%%%%%%%%%%%%%%%%%%%%%%%% END table3
	%%%%%%%%%%%%%%%%%%%%%%%%%%% Time consumption
	
	\begin{table}[h]
		\large
		\caption{The running time of each combination, each cell means time consumption $(s)$ when feeding 10/100/1000 batch size of data.}
		\renewcommand\arraystretch{1.7}
		\centering
		%\begin{center}
		\resizebox{\textwidth}{!}{\begin{tabular}{|p{2cm}<{\centering}|c|c|c|c|c|c|}
				\hline  %添加表格头部粗线
				& \textbf{VGG}& \textbf{ResNet}& \textbf{DDCResNet}& \textbf{DWT}& \textbf{SVD}& \textbf{FFT}\\
				%\midrule  %添加表格中横线
				\hline
				\textbf{Adaboost}& 3.213/3.959/19.677& 3.776/4.213/20.581& 6.456/6.86/26.100& 0.184/2.652/18.693& 0.165/1.786/14.445& 0.130/3.994/280.162\\
				\hline
				\textbf{SVM$^{1}$}& 3.176/3.905/16.530& 3.739/4.159/17.434& 6.419/6.806/22.953& 0.160/2.013/15.674& 0.147/1.292/13.010& 0.184/3.207/48.925\\
				\hline
				\textbf{SVM$^{2}$}& 3.176/3.905/16.960& 3.739/4.159/17.864& 6.419/6.806/23.383& 0.143/2.412/15.108& 0.153/1.292/12.967& 0.029/1.688/147.075\\
				\hline
				\textbf{SVM$^{3}$}& 3.176/3.905/16.902& 3.739/4.159/17.806& 6.419/6.806/23.325& 0.296/1.609/16.883& \textbf{0.146/1.291/12.944}& 0.028/1.665/143.031\\
				\hline
				\textbf{KNN}& 3.180/3.904/16.602& 3.743/4.158/17.506& 6.423/6.805/23.025& 0.198/1.059/18.370& 0.162/1.294/12.889& 0.030/1.393/89.841\\
				\hline
				\textbf{J48}& 3.184/3.927/16.331& 3.747/4.181/17.235& 6.427/6.828/22.754& 0.184/1.477/15.854& 0.148/1.296/12.791& 0.858/4.822/201.731\\
				\hline
				\textbf{CNN}& 3.189/3.903/15.963& 3.752/4.157/16.867& 6.432/6.804/22.386& 0.139/1.749/17.110& 0.156/1.300/12.756& 1.200/4.190/341.870\\
				\hline
				\textbf{ID3}& 3.176/3.920/16.629& 3.739/4.174/17.533& 6.419/6.821/23.052& 0.142/1.842/15.556& 0.1467/1.291/12.897& 0.062/1.756/43.519\\
				\hline
				\textbf{LDA}& 3.176/3.901/16.292& 3.739/4.154/17.196& 6.419/6.801/22.715& 0.148/1.466/16.770& 0.147/1.292/12.967& 0.138/2.478/41.830\\
				\hline %添加表格底部粗线
		\end{tabular}}
		\label{TimeMeasurement}
		%\end{center}
	\end{table}
	
	\begin{figure}[h!]
		\centering %图片居中
		\includegraphics[width=0.6\textwidth]{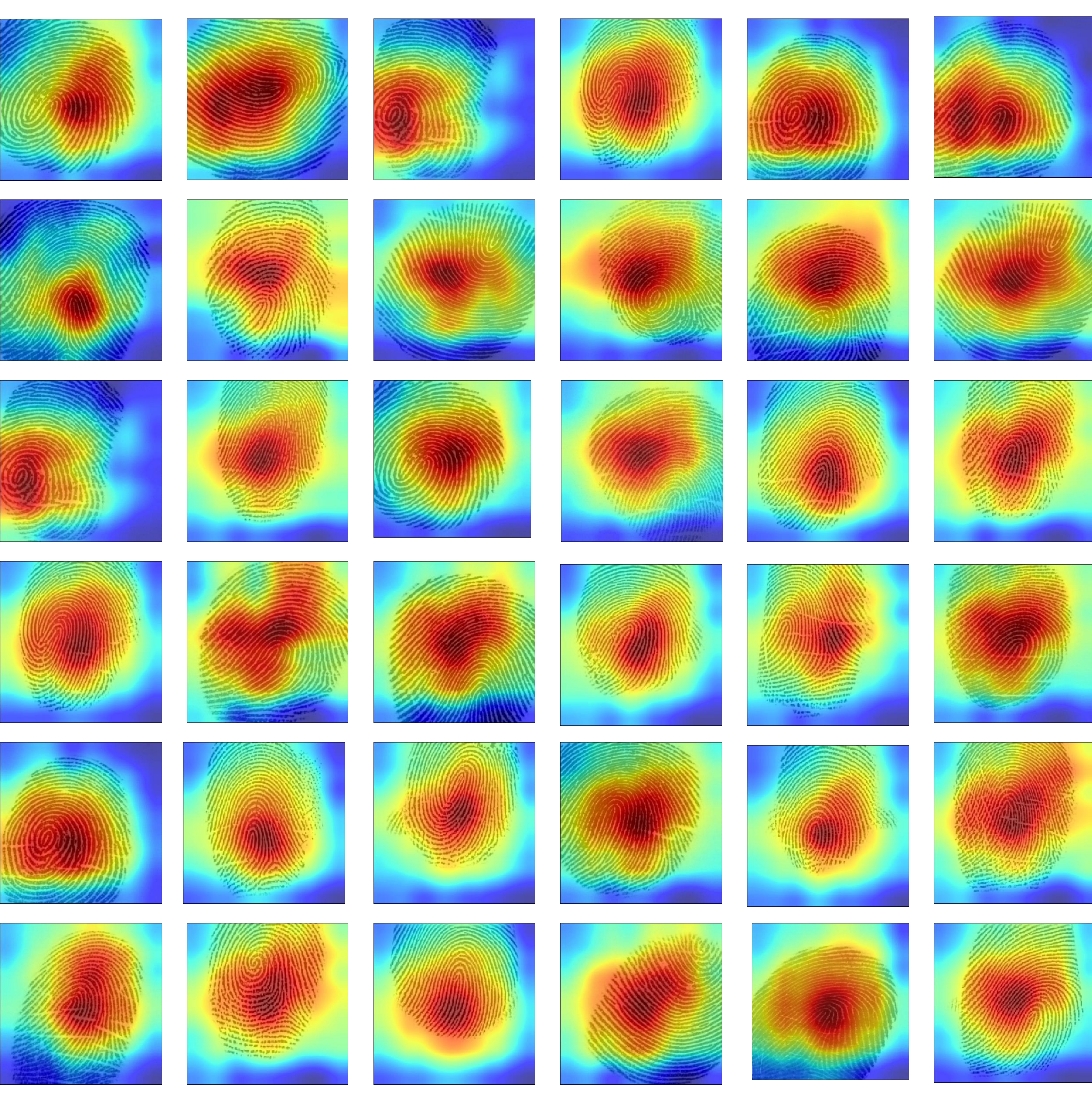}
		\caption{Grad-CAM visualizations for DDC-ResNet. The red regions in the figures correspond to high score for gender information. Best viewed in color.} %最终文档中希望显示的图片标题
		\label{fig:fig4}
	\end{figure}
	
	%%%%%%%%%%%%%%%%%%%%%%%%%%% END Time consumption 
	\subsection{Effect of Features}
	To further research on gender-related features, we apply the grad-cam~\cite{selvaraju2017grad} technique to visualize the heat maps, seeing which part contributes most. Specifically, based on earlier results, we visualize testing fingers by using the DDC-ResNet extractor. The overview of visualization is exhibited in Figure 5, It can be seen that concentrations are mainly on or around the center of fingerprints. Detailed observation is shown in Figure 6. The concentration covers the whorl part which belongs to the pattern features of level 1. Further observation reveals that bifurcations are gathered in concentration regions. However, we exclude the sweat pores around the bifurcations since other parts containing more remarkable sweat pores have not received attention. We also find that the concentration regions prefer to be continuous parts. Therefore, it is not hard to say that the foundation of recognition is complete ridges, which explains the reason for manual extraction of minutiae features in well-selected continuous regions.\\
	
	%---------------------------gradcams 

	\begin{figure}[h!]
		\centering %图片居中
		\includegraphics[width=0.7\textwidth]{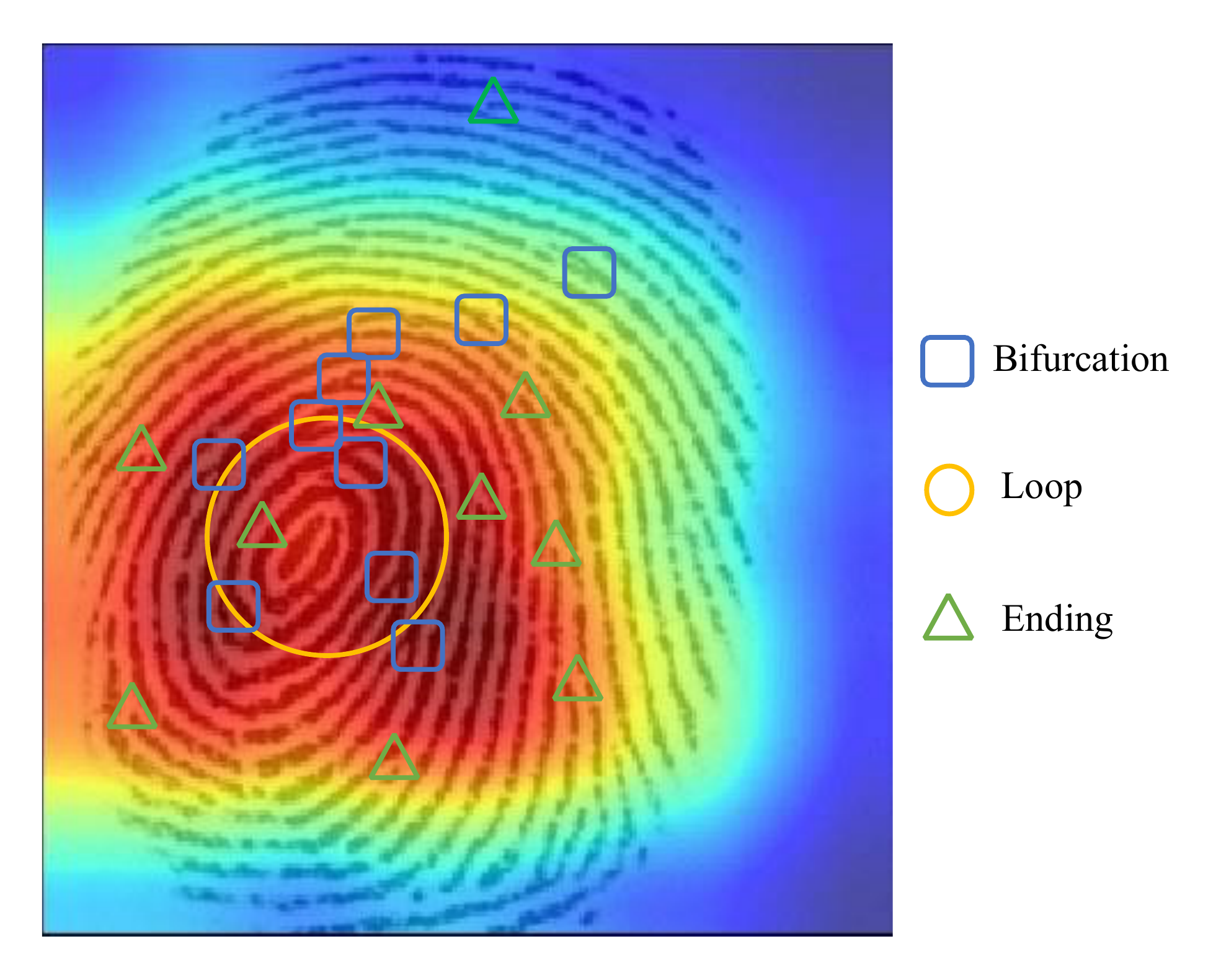}
		\caption{Visualizations of bifurcation, loop and ending in a fingerprint. The red region with more gender information in the center of the fingerprint gets higher scores.} %最终文档中希望显示的图片标题
		\label{fig:fig4}
	\end{figure}
	
	\section{Conclusion and Future work}
	This paper proposes an effective network considering the global reception field in the gender classification task, which is realized by replacing normal convolutions with dilated convolution in the extraction method. The experiment thoroughly explores the efficiency from three aspects. First, comparing our method with various methods with fair implementation details in our dataset. 6 typical automatic feature extraction methods like DWT, SVD, VGG, and our method coupling with 9 mainstream classifiers such as Adaboost, kNN, SVM, CNN, etc. are evaluated. Experimental results reveal that the combination of our extractor with the CNN classifier outperforms other combinations. For average accuracy, it reaches 96.50\% and for separate-gender accuracy, it reaches 0.9752 (males) / 0.9548 (females). Second, we investigate the effect of fingers by classifying gender using separate fingers, and find the best-performing finger is the right ring finger, which reaches an accuracy of 92.455\%. Third, we study the effect of features by visualizing concentrations of fingerprints. Depending on the analysis, loops/whorls (level 1), bifurcations (level 2) and line shapes (level 3) may have a close relationship with gender. This work not only comprehensively explores the efficiency of the proposed method, but also provides a way to observe fingerprint identification features much closer. These specific features will be quantified in the future to further explore the impact on fingerprint recognition.

	\bibliographystyle{unsrtnat}
	\bibliography{references}  %%% Uncomment this line and comment out the ``thebibliography'' section below to use the external .bib file (using bibtex) .

\begin{thebibliography}{28}
\providecommand{\natexlab}[1]{#1}
\providecommand{\url}[1]{\texttt{#1}}
\expandafter\ifx\csname urlstyle\endcsname\relax
  \providecommand{\doi}[1]{doi: #1}\else
  \providecommand{\doi}{doi: \begingroup \urlstyle{rm}\Url}\fi

\bibitem[Abdullah et~al.(2016{\natexlab{a}})Abdullah, Rahman, Abas, and
  Saad]{abdullah2016fingerprint}
SF~Abdullah, AFNA Rahman, ZA~Abas, and WHM Saad.
\newblock Fingerprint gender classification using univariate decision tree
  (j48).
\newblock \emph{Network (MLPNN)}, 96\penalty0 (95.27):\penalty0 95--95,
  2016{\natexlab{a}}.

\bibitem[Abdullah et~al.(2016{\natexlab{b}})Abdullah, Rahman, Abas, and
  Saad]{abdullah2016support}
Siti~Fairuz Abdullah, AFNA Rahman, ZA~Abas, and WHM Saad.
\newblock Support vector machine, multilayer perceptron neural network, bayes
  net and k-nearest neighbor in classifying gender using fingerprint features.
\newblock \emph{International Journal of Computer Science and Information
  Security}, 14\penalty0 (7):\penalty0 336, 2016{\natexlab{b}}.

\bibitem[Gnanasivam and Muttan(2012)]{gnanasivam2012fingerprint}
P~Gnanasivam and Dr~S Muttan.
\newblock Fingerprint gender classification using wavelet transform and
  singular value decomposition.
\newblock \emph{arXiv preprint arXiv:1205.6745}, 2012.

\bibitem[Gupta and Rao(2014)]{gupta2014fingerprint}
Samta Gupta and A~Prabhakar Rao.
\newblock Fingerprint based gender classification using discrete wavelet
  transform \& artificial neural network.
\newblock \emph{International Journal of Computer Science and mobile
  computing}, 3\penalty0 (4):\penalty0 1289--1296, 2014.

\bibitem[Mishra and Maheshwary(2017)]{mishra2017novel}
Ashish Mishra and Preeti Maheshwary.
\newblock A novel technique for fingerprint classification based on naive bayes
  classifier and support vector machine.
\newblock \emph{International Journal of Computer Applications}, 975:\penalty0
  8887, 2017.

\bibitem[Rekha et~al.(2019)Rekha, Gurupriya, Gayadhri, and
  Sowmya]{rekha2019dactyloscopy}
V~Rekha, S~Gurupriya, S~Gayadhri, and S~Sowmya.
\newblock Dactyloscopy based gender classification using machine learning.
\newblock In \emph{2019 IEEE International Conference on System, Computation,
  Automation and Networking (ICSCAN)}, pages 1--5. IEEE, 2019.

\bibitem[Shinde and Annadate(2015)]{shinde2015analysis}
Mangesh~K Shinde and SA~Annadate.
\newblock Analysis of fingerprint image for gender classification or
  identification: using wavelet transform and singular value decomposition.
\newblock In \emph{2015 International Conference on Computing Communication
  Control and Automation}, pages 650--654. IEEE, 2015.

\bibitem[Wedpathak et~al.(2018)Wedpathak, Kadam, Kadam, Mhetre, and
  Jankar]{wedpathak2018fingerprint}
Ganesh~S Wedpathak, DG~Kadam, KG~Kadam, AR~Mhetre, and VK~Jankar.
\newblock Fingerprint based gender classification using ann.
\newblock \emph{International Journal of Recent Trends in Engineering \&
  Research (IJRTER)}, 4\penalty0 (3):\penalty0 4, 2018.

\bibitem[Karu and Jain(1996)]{karu1996fingerprint}
Kalle Karu and Anil~K Jain.
\newblock Fingerprint classification.
\newblock \emph{Pattern recognition}, 29\penalty0 (3):\penalty0 389--404, 1996.

\bibitem[Henry(1913)]{henry1913classification}
Edward~Richard Henry.
\newblock \emph{Classification and uses of finger prints}.
\newblock HM Stationery Office, printed by Darling and son, Limited, 1913.

\bibitem[Arun and Sarath(2011)]{arun2011machine}
KS~Arun and KS~Sarath.
\newblock A machine learning approach for fingerprint based gender
  identification.
\newblock In \emph{2011 IEEE Recent Advances in Intelligent Computational
  Systems}, pages 163--167. IEEE, 2011.

\bibitem[Badawi et~al.(2006)Badawi, Mahfouz, Tadross, and
  Jantz]{badawi2006fingerprint}
Ahmed~M Badawi, Mohamed Mahfouz, Rimon Tadross, and Richard Jantz.
\newblock Fingerprint-based gender classification.
\newblock \emph{IPCV}, 6\penalty0 (8):\penalty0 l, 2006.

\bibitem[Kralik and Novotny(2003)]{kralik2003epidermal}
Miroslav Kralik and Vladimir Novotny.
\newblock Epidermal ridge breadth: an indicator of age and sex in
  paleodermatoglyphics.
\newblock \emph{Variability and evolution}, 11\penalty0 (2003):\penalty0 5--30,
  2003.

\bibitem[Kaur and Mazumdar(2012)]{kaur2012fingerprint}
Ritu Kaur and Susmita~Ghosh Mazumdar.
\newblock Fingerprint based gender identification using frequency domain
  analysis.
\newblock \emph{International Journal of Advances in Engineering \&
  Technology}, 3\penalty0 (1):\penalty0 295, 2012.

\bibitem[Marasco et~al.(2014)Marasco, Lugini, and Cukic]{marasco2014exploiting}
Emanuela Marasco, Luca Lugini, and Bojan Cukic.
\newblock Exploiting quality and texture features to estimate age and gender
  from fingerprints.
\newblock In \emph{Biometric and Surveillance Technology for Human and Activity
  Identification XI}, volume 9075, page 90750F. International Society for
  Optics and Photonics, 2014.

\bibitem[Chen et~al.(2017)Chen, Shi, Zhang, Wu, and Guizani]{chen2017deep}
Min Chen, Xiaobo Shi, Yin Zhang, Di~Wu, and Mohsen Guizani.
\newblock Deep features learning for medical image analysis with convolutional
  autoencoder neural network.
\newblock \emph{IEEE Transactions on Big Data}, 2017.

\bibitem[Acree(1999)]{acree1999there}
Mark~A Acree.
\newblock Is there a gender difference in fingerprint ridge density?
\newblock \emph{Forensic science international}, 102\penalty0 (1):\penalty0
  35--44, 1999.

\bibitem[Sudesh~Gungadin(2007)]{gungadin2007sex}
MBBS Sudesh~Gungadin.
\newblock Sex determination from fingerprint ridge density.
\newblock \emph{Internet Journal of Medical Update}, 2\penalty0 (2), 2007.

\bibitem[Abdullah et~al.(2016{\natexlab{c}})Abdullah, Rahman, Abas, and
  Saad]{abdullah2016multilayer}
SF~Abdullah, AFNA Rahman, ZA~Abas, and WHM Saad.
\newblock Multilayer perceptron neural network in classifying gender using
  fingerprint global level features.
\newblock \emph{Indian Journal of Science and Technology}, 9\penalty0
  (9):\penalty0 1--6, 2016{\natexlab{c}}.

\bibitem[Sheetlani et~al.(2017)Sheetlani, Pardeshi,
  et~al.]{sheetlani2017fingerprint}
Jitendra Sheetlani, Rajmohan Pardeshi, et~al.
\newblock Fingerprint based automatic human gender identification.
\newblock \emph{Int. J. Comput. Appl}, 170\penalty0 (7):\penalty0 1--4, 2017.

\bibitem[Alam et~al.(2019)Alam, Dua, Gupta, et~al.]{alam2019comparative}
Shadab Alam, Megha Dua, Ashutosh Gupta, et~al.
\newblock A comparative study of gender classification using fingerprints.
\newblock In \emph{2019 6th International Conference on Computing for
  Sustainable Global Development (INDIACom)}, pages 880--884. IEEE, 2019.

\bibitem[Wang et~al.(2018)Wang, Chen, Yuan, Liu, Huang, Hou, and
  Cottrell]{wang2018understanding}
Panqu Wang, Pengfei Chen, Ye~Yuan, Ding Liu, Zehua Huang, Xiaodi Hou, and
  Garrison Cottrell.
\newblock Understanding convolution for semantic segmentation.
\newblock In \emph{2018 IEEE winter conference on applications of computer
  vision (WACV)}, pages 1451--1460. IEEE, 2018.

\bibitem[Maas et~al.(2013)Maas, Hannun, Ng, et~al.]{maas2013rectifier}
Andrew~L Maas, Awni~Y Hannun, Andrew~Y Ng, et~al.
\newblock Rectifier nonlinearities improve neural network acoustic models.
\newblock In \emph{Proc. icml}, volume~30, page~3. Citeseer, 2013.

\bibitem[Abadi et~al.(2016)Abadi, Agarwal, Barham, Brevdo, Chen, Citro,
  Corrado, Davis, Dean, Devin, et~al.]{abadi2016tensorflow}
Mart{\'\i}n Abadi, Ashish Agarwal, Paul Barham, Eugene Brevdo, Zhifeng Chen,
  Craig Citro, Greg~S Corrado, Andy Davis, Jeffrey Dean, Matthieu Devin, et~al.
\newblock Tensorflow: Large-scale machine learning on heterogeneous distributed
  systems.
\newblock \emph{arXiv preprint arXiv:1603.04467}, 2016.

\bibitem[Holmes et~al.(1994)Holmes, Donkin, and Witten]{holmes1994weka}
Geoffrey Holmes, Andrew Donkin, and Ian~H Witten.
\newblock Weka: A machine learning workbench.
\newblock In \emph{Proceedings of ANZIIS'94-Australian New Zealnd Intelligent
  Information Systems Conference}, pages 357--361. IEEE, 1994.

\bibitem[Pedregosa et~al.(2011)Pedregosa, Varoquaux, Gramfort, Michel, Thirion,
  Grisel, Blondel, Prettenhofer, Weiss, Dubourg, et~al.]{pedregosa2011scikit}
Fabian Pedregosa, Ga{\"e}l Varoquaux, Alexandre Gramfort, Vincent Michel,
  Bertrand Thirion, Olivier Grisel, Mathieu Blondel, Peter Prettenhofer, Ron
  Weiss, Vincent Dubourg, et~al.
\newblock Scikit-learn: Machine learning in python.
\newblock \emph{the Journal of machine Learning research}, 12:\penalty0
  2825--2830, 2011.

\bibitem[Gnanasivam and Vijayarajan(2019)]{gnanasivam2019gender}
P~Gnanasivam and R~Vijayarajan.
\newblock Gender classification from fingerprint ridge count and fingertip size
  using optimal score assignment.
\newblock \emph{Complex \& Intelligent Systems}, 5\penalty0 (3):\penalty0
  343--352, 2019.

\bibitem[Selvaraju et~al.(2017)Selvaraju, Cogswell, Das, Vedantam, Parikh, and
  Batra]{selvaraju2017grad}
Ramprasaath~R Selvaraju, Michael Cogswell, Abhishek Das, Ramakrishna Vedantam,
  Devi Parikh, and Dhruv Batra.
\newblock Grad-cam: Visual explanations from deep networks via gradient-based
  localization.
\newblock In \emph{Proceedings of the IEEE international conference on computer
  vision}, pages 618--626, 2017.

\end{thebibliography}

	%%% Uncomment this section and comment out the \bibliography{references} line above to use inline references.
	% \begin{thebibliography}{1}
	
	% 	\bibitem{kour2014real}
	% 	George Kour and Raid Saabne.
	% 	\newblock Real-time segmentation of on-line handwritten arabic script.
	% 	\newblock In {\em Frontiers in Handwriting Recognition (ICFHR), 2014 14th
	% 			International Conference on}, pages 417--422. IEEE, 2014.
	
	% 	\bibitem{kour2014fast}
	% 	George Kour and Raid Saabne.
	% 	\newblock Fast classification of handwritten on-line arabic characters.
	% 	\newblock In {\em Soft Computing and Pattern Recognition (SoCPaR), 2014 6th
	% 			International Conference of}, pages 312--318. IEEE, 2014.
	
	% 	\bibitem{hadash2018estimate}
	% 	Guy Hadash, Einat Kermany, Boaz Carmeli, Ofer Lavi, George Kour, and Alon
	% 	Jacovi.
	% 	\newblock Estimate and replace: A novel approach to integrating deep neural
	% 	networks with existing applications.
	% 	\newblock {\em arXiv preprint arXiv:1804.09028}, 2018.
	
	% \end{thebibliography}

\end{document}